# Validation of computer simulations of the HyQ robot

Marco Frigerio, Victor Barasuol, Michele Focchi and Claudio Semini *
Department of Advanced Robotics, Istituto Italiano di Tecnologia
Genova, Italy

April 22, 2016

## 1 Introduction

This short technical report illustrates the results of a test procedure we performed to validate the computer simulation of the HyQ robot.

## 2 The HyQ robot

Fig. 1 shows a picture of HyQ, a quadruped robot with hydraulically actuated joints [Sem10; Sem11]. The machine weighs 80 kg, is roughly 1

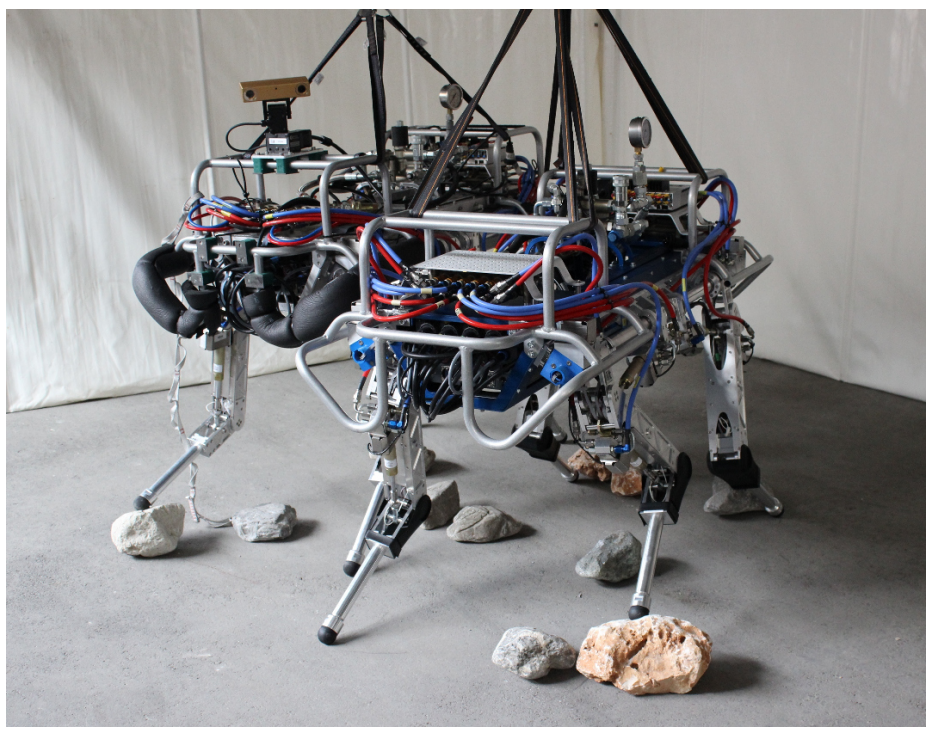

**Figure 1:** *The HyQ robot*

meter long and has a leg length of 0.78 m with fully-extended legs. All of its 12 degrees of freedom (DOF) are torque-controlled joints: The hip abduction/adduction (HAA) joints are driven by rotary hydraulic actuators with strain-gauge based torque sensors for torque control. All 8 joints in the sagittal plane (hip flexion/extension (HFE) and knee flexion/extension (KFE)) are actuated by hydraulic cylinders, that are connected to load cells for force measurement. High-performance servovalves (MOOG E024) enable joint-level torque control with excellent tracking that led to the implementation of active impedance [Boa12; Sem15].

| Property/Feature | Value |
|---|---|
| dimensions | 1.0 x 0.5 x 0.98 m (LxWxH) |
| approximate leg length (fully extended) | 0.78 m |
| weight | 80kg |
| active DOF | 12 |
| hydraulic actuation) | double–vane rotary actuators and double–acting asymmetric cylinders |
| motion range | 90° (HAA), 120° (HFE, KFE) |
| max joint torque (HAA) | 120Nm (peak torque at 20MPa) |
| max joint torque (HFE, KFE) | 181Nm (peak torque at 20MPa) |
| position sensors | relative encoder, 80000CPR |
| torque sensors | custom torque sensor (HAA), 5kN loadcell (HFE, KFE) |
| perception sensors | IMU, stereo camera, lidar |
| onboard computer | Intel i5 based computer, 8GB of RAM |
| joint control (rate) | position and torque (1kHz) |
| locomotion skills | walking (crawl, trot), running (flying trot), hopping, squat jumping, rearing |

**Table 1:** *Overview of the specifications and features of the HyQ robot. HAA stands for Hip Abduction Adduction, HFE for Hip Flexion Extension, KFE for Knee Flexion Extension.*

Table 1 lists the main specifications and features of the robot.

# 3 Simulation environment

The *SL simulator and motor controller package* was adopted as the first software control system of HyQ. SL provides a hard real–time compatible infrastructure for motor control (i.e. low level control of the joint motion/force) and for trajectory generation [Sch09]. SL was entirely written in C/C++, and can be built and executed on top of a real–time Xenomai–based Linux system.

Thanks to a general I/O interface, the motor control is agnostic to the actual component that would be receiving the control commands. There-fore, with SL, the control software can be deployed either on a computer connected with the real hardware of a robot, as well as on a computer running a physics simulator of the same robot. No significant change of code is required between the two options, which guarantees that simulations and real experiments are driven by the same software. The SL package itself includes a physics simulator, which is the one we used.

The kinematics and dynamics engines for the HyQ robot (e.g. the forward dynamics routine required by the simulator) have been implemented using RobCoGen, a generator of robot–specific, optimized code for the most common kinematics and dynamics routines used in robotics [Fri16]. RobCoGen was not validated by any formal method, but the generated code was tested by numerical comparisons with other libraries/tools, for a variety of articulated robot models.

# 4 Experimental Validation of the Simulator

To validate the simulation environment we performed 1m/s walking trot experiments and compared it with the results of the same behaviour in simulation. To generate a stable trotting motion we used the locomotion controller presented in [Bar13].

Figure 2 shows the knee torque profiles of all the legs, in the case of both simulation and real experiments. The plots in the bottom show the power obtained by multiplying the joint torque with the joint velocity. The negative power regions represent periods in the gait cycle where energy is injected into the joint, e.g. at foot touch down in the beginning of each stance phase.

Note that the torque peaks during foot touch down in the left front leg are only present in the experimental data. These peaks are created by an imperfect trotting motion. Such disturbances may be due to inaccurate state estimation, or imperfections in the dynamics model of the robot.

Figure 3 shows the position and velocity profiles for the same joint during the same motion. All the plots show a sufficient similarity between simulation and real experiment. On the basis of these results we can use our simulation environment to effectively test new behaviours, and to aid the design of new robots.

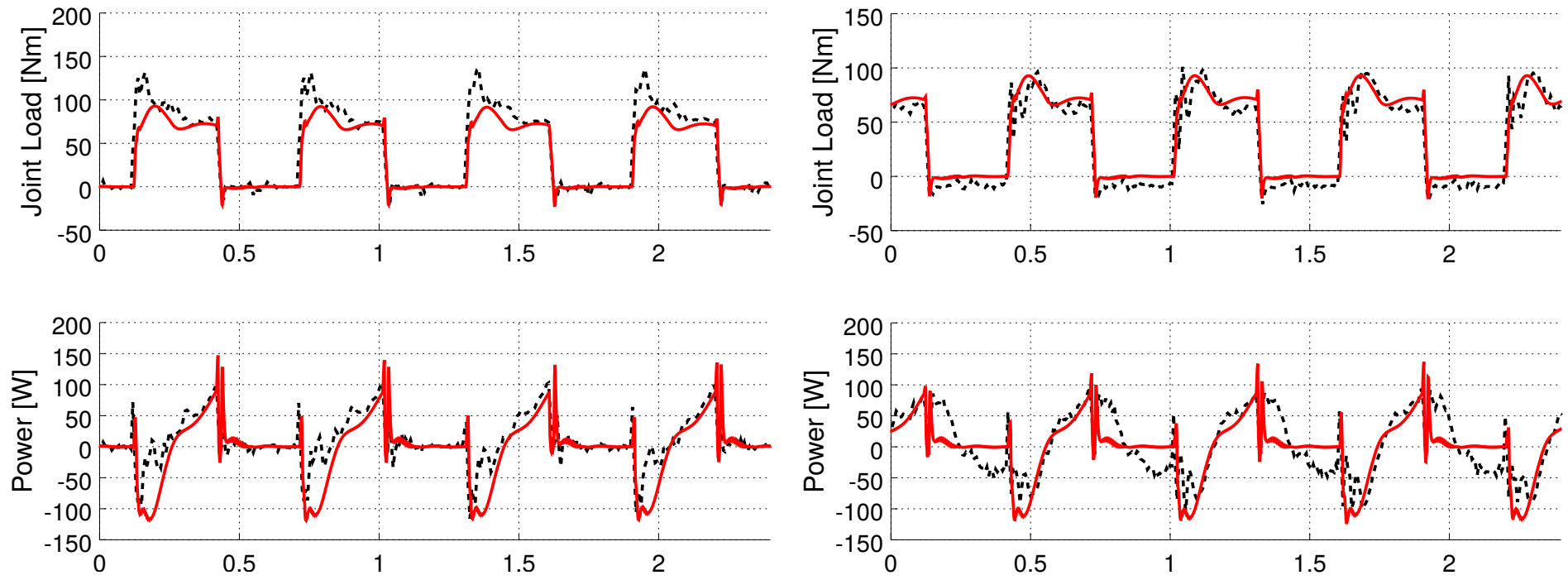

**(a)** *Left Front leg* **(b)** *Right Front leg*

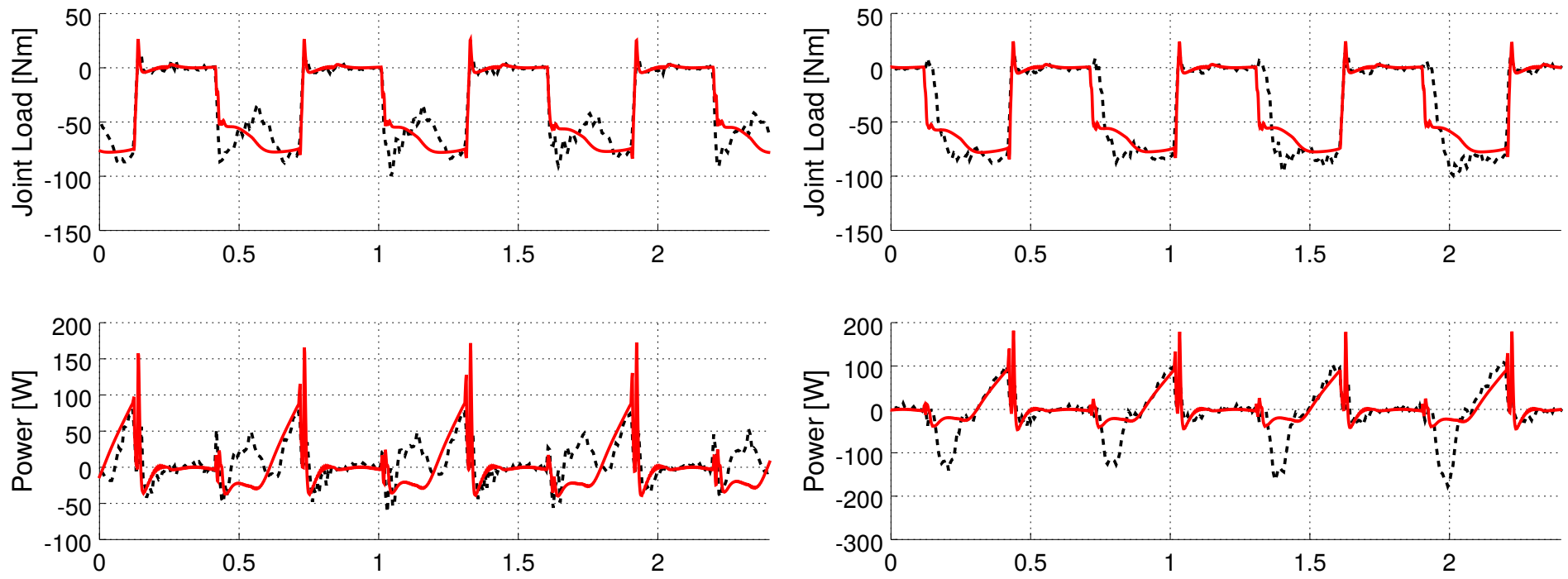

**(c)** *Left Hind leg* **(d)** *Right Hind leg*

**Figure 2:** *Comparison between simulation (red solid) and experimental (black dashed) results of a 1.0 m/s walking trot with the HyQ robot. The bottom plots illustrate the mechanical power profile obtained by multiplying the joint torque with the joint velocity. All plots refer to the knee joint. The negative power areas indicate periods in which energy is injected into the joint.*

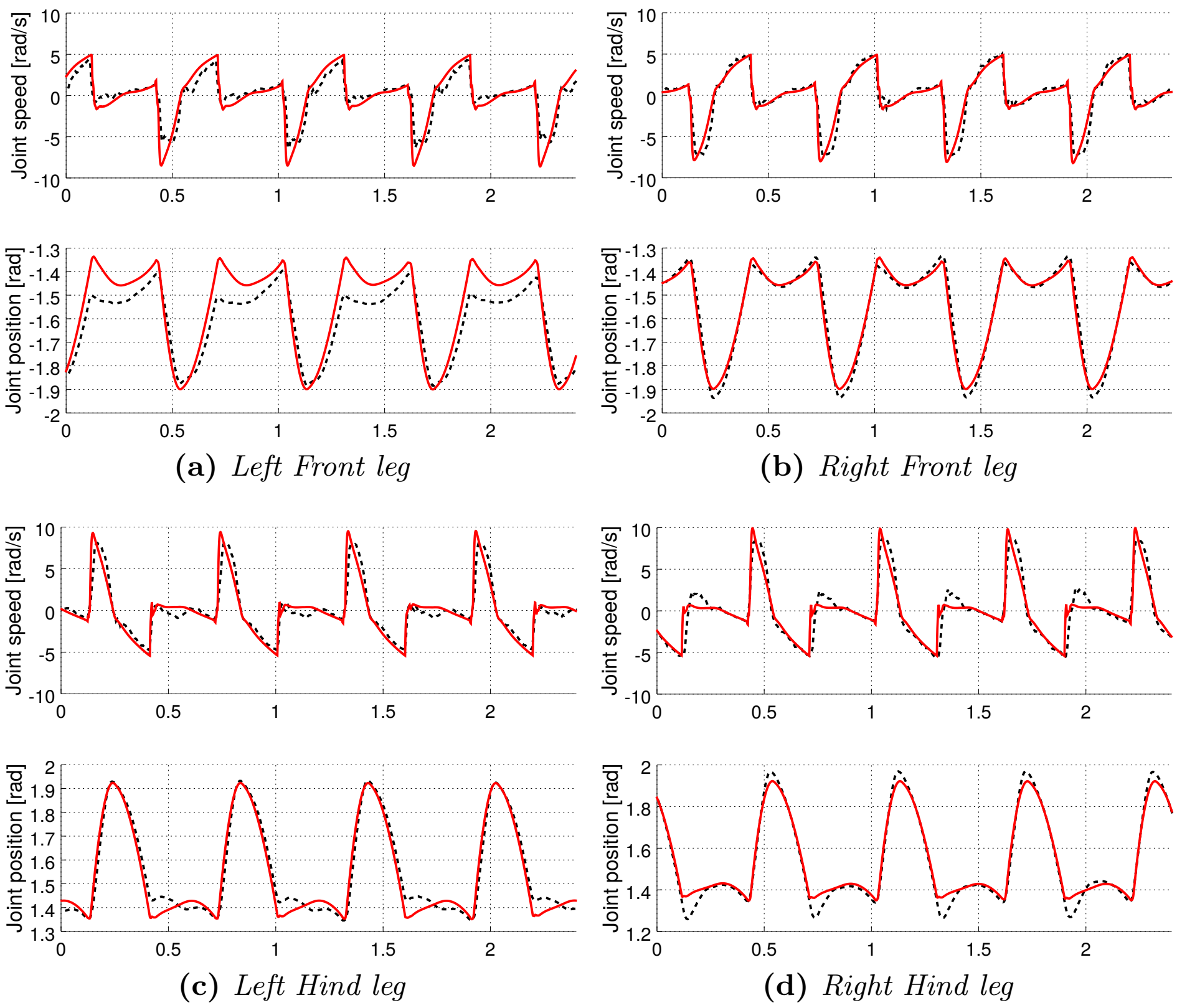

**Figure 3:** *Comparison between simulation (red solid) and experimental (black dashed) results of a 1.0 m/s walking trot with the HyQ robot; see Figure 2. The bottom plots show the knee joint position, whereas the top ones show the velocity.*